\documentclass[runningheads]{llncs}
\usepackage{graphicx}

\usepackage{tikz}
\usepackage{comment}
\usepackage{amsmath,amssymb, amsfonts} 
\usepackage{ctable}
\usepackage{color}
\usepackage{subcaption}
\usepackage{ragged2e}
\usepackage{multirow}
\usepackage{cite}

\usepackage[accsupp]{axessibility}  

\begin{document}
\pagestyle{headings}
\mainmatter
\def\ECCVSubNumber{11}  

\title{An Evaluation of Self-Supervised Pre-Training for Skin-Lesion Analysis} 

\titlerunning{An Evaluation of Self-Supervised Pre-Training for Skin-Lesion Analysis}
%
\author{Levy Chaves\inst{1,3}\orcidID{0000-0002-7431-2440} \and
Alceu Bissoto\inst{1,3}\orcidID{0000-0003-2293-6160} \and
Eduardo Valle\inst{2,3}\orcidID{0000-0001-5396-9868} \and Sandra Avila\inst{1,3}\orcidID{0000-0001-9068-938X}}
\authorrunning{L. Chaves et al.}
%
\institute{Institute of Computing (IC) \\ \email{\{levy.chaves, alceubissoto, sandra\}@ic.unicamp.br} \and
School of Electrical and Computing Engineering (FEEC) \\ 
\email{dovalle@dca.fee.unicamp.br} \\
\and
Recod.ai Lab., University of Campinas
}
\maketitle

\begin{abstract}
Self-supervised pre-training appears as an advantageous alternative to supervised pre-trained for transfer learning. By synthesizing annotations on pretext tasks, self-supervision allows pre-training models on large amounts of pseudo-labels before fine-tuning them on the target task. In this work, we assess self-supervision for diagnosing skin lesions, comparing three self-supervised pipelines to a challenging supervised baseline, on five test datasets comprising in- and out-of-distribution samples. Our results show that self-supervision is competitive both in improving accuracies and in reducing the variability of outcomes. Self-supervision proves particularly useful for low training data scenarios ($<$1500 and $<$150 samples), where its ability to stabilize the outcomes is essential to provide sound results. 

\keywords{Self-supervision, Out-of-distribution, Skin lesions, Melanoma, Classification, Small Datasets}
\end{abstract}

\section{Introduction}

Self-supervised learning bridges the gap between supervised learning, which leads to the most accurate models but requires human-annotated samples, and unsupervised learning, which can exploit non-annotated samples but often leads to disappointing accuracies. By using synthesized annotations on so-called \textit{pretext tasks}, self-supervision is able to \textit{pre-train} models on abundant pseudo-labels before tuning them for the downstream target task.

Applications for which annotated data is expensive or scarce — often the case for medical applications — especially benefit from self-supervision~\cite{vu2021medaug,azizi2021big,moris-2021-eye}. Training state-of-the-art Deep Learning models require extensive training datasets, which are seldom available for medical applications. We can mitigate the issue by applying transfer learning, i.e., pre-training the models (with classical supervised learning) on a large, unrelated dataset, and fine-tuning them on the target dataset, but there is a risk that the representations learned during pre-training will not fully adapt to the downstream task~\cite{menegola2017knowledge}. Self-supervised pre-training has proved, thus, advantageous for transfer learning in many tasks, such as object localization~\cite{NEURIPS2020_7288251b}, speech representation~\cite{kawakami2020learning}, and medical image classification~\cite{surpassimagenet,truong2021transferable,azizi2022robust}. 

In this work, we assess self-supervision pre-training for the automated diagnosis of skin lesions, an application for which traditionally transfer learning from models supervised on ImageNet is employed to mitigate the scarcity of data~\cite{menegola2017knowledge,VALLE2020303}. 
Our work improves on existing self-supervised applications to medical applications~\cite{truong2021transferable,azizi2021big,hosseinzadeh2021systematic,verdelho-isbi-2022} by evaluating performances on out-of-distribution and low data (less than 150 samples in the stringiest case) regimens. We also evaluate adding an intermediate contrastive learning pre-training before performing the traditional fine-tuning protocol. 

The main contributions of this work are:

\begin{itemize}
    \item  We assess five self-supervision learning candidates (BYOL, 
    InfoMin, 
    MoCo, 
    SimCLR, 
    and SwAV) 
    against a competitive supervised baseline;
    \item We perform a systematic assessment of four transfer learning pipelines (the supervised baseline and three self-supervised contenders) in five publicly accessible test datasets, comprising in-distribution and out-distribution scenarios. Our results suggest that self-supervised models present superior performance in both in-distribution and out-of-distribution in almost all evaluated datasets;
    \item We assess the performance of our pipelines/datasets in a low-data training scenario (with as few as $148$ samples). Again, we find performance improvement in favor of self-supervised pre-training. 
\end{itemize}

We organized the remaining text as follows. We discuss the state-of-the-art on self-supervision in Sec.~\ref{sec:relatedwork}, comprising both general works and those dedicated to medical images, and skin lesions in particular. We detail our goals, datasets, pipelines, protocols, experimental design, and implementation details in Sec.~\ref{sec:methods}. Experimental results and analyses appear in Sec.~\ref{sec:results}. Finally, we discuss our main findings, along with future research directions in Sec.~\ref{sec:conclusion}.


\section{Related Work}
\label{sec:relatedwork}

Self-supervised learning has attracted growing attention in the past decade, with hundreds of papers published in the past few years. For a comprehensive review, we refer the reader to the survey of Jing et al.~\cite{jing2020self}. We recommend the survey by Liu et al.~\cite{liu-generative-contrastive} for a more fundamental/theoretical viewpoint on a broad scope of techniques. In this section, we will limit ourselves to the methods directly relevant to this work, and to a selection of methods used for medical images and, in particular, for skin-lesion analysis.

\subsection{Self-supervised Learning for Visual Tasks}
\label{sec:related-sslgeneral}

Self-supervised learning pre-trains models on auxiliary \textbf{pretext tasks} such as colorizing~\cite{zhang2016colorful}, predicting rotation angles~\cite{gidaris2018unsupervised}, and in-painting~\cite{pathak2016context}, before fine-tuning them on the \textbf{downstream task}, i.e., the \textbf{target task}. This allows pre-learning representations on unlabeled data and then refining those representations on labeled data. The base model in self-supervised learning, called the \textbf{encoder}, transforms the input image into the \textbf{(latent) representations}. ResNet-50 is often employed as a backbone, due to its ability to conciliating simplicity and accuracy~\cite{Wu_2018_CVPR,chen2020simple,moco,byol,caron2020unsupervised,azizi2021big}.

A critical breakthrough in self-supervised learning was the adoption of contrastive losses~\cite{NEURIPS2020_8965f766,jing2020self}, which explicitly organize the feature space by bringing together the representations for related (positive) pairs of samples, while pushing apart the representations for unrelated (negative) pairs. 

InstDisc~\cite{Wu_2018_CVPR} is the seminal work on contrastive self-supervised image classification. InstDisc reframes class-level classification as instance-level discrimination: each training sample becomes one label, whose data-augmented views must be recognized against data-augmented views from all other training samples. The challenge is extending the loss for so many labels (millions, in ImageNet), which is conquered by reformulating the softmax loss. An $\ell_2$-normalization turns the dot products into cosine similarities. A temperature hyperparameter $\tau$ allows regulating the loss concentration. A memory bank caches the parameters for each label/instance. Finally, the softmax is approximated using the Noise-Contrastive Estimation~\cite{gutmann2010noise}, previously successful for training very large word embeddings. The technique creates very compact (\text{$128$-d}) representations, thus making storage and computation for the memory bank feasible, despite the large number of entries.
 
Instead of using a memory bank, SimCLR~\cite{chen2020simple}  employs end-to-end learning~\cite{moco}, adding an auxiliary dimension-reducing network (projection network) after the encoder and generating the representations on the fly for each batch. The pretext task and loss are very similar to InstDisc's, but only the samples present in the batch are considered in the computation of the loss, without resorting to the memory bank. Thus, SimCLR requires very large batch sizes ($4096$ samples \textit{vs.} InstDisc' $256$), and strong data-augmented views in order to be effective.

MoCo~\cite{moco} proposes a dictionary of representations whose size is a hyperparameter that may be much larger than the batch size (which is limited to the GPU memory) while still being much smaller than the training set as in InstDisc. The entries in the dictionary are the past few batches, updated in a FIFO scheme. Pretext task and loss still work similar to InstDisc and SimCLR, but negative examples are now taken from the dictionary. The parameters for each label are updated using a “momentum” update, which prevents the representations from fluctuating too much. That is reminiscent of the proximal regularization of InstDisc, but the latter acts on the loss instead of the representations. Further, MoCo-V2~\cite{chen2020improved} added the projection network and strong data-augmented views as in SimCLR's into the original MoCo formulation. 

In BYOL~\cite{byol}, one slow network creates targets for a fast network. The parameters of the fast network are learned by backpropagation, and the parameters of the slow network are the exponential moving average of the parameters of the fast network. In that manner, BYOL bootstraps its own target representations. BYOL still matches data-augmented views between positive pairs as pretext, but without resorting to negative pairs. Instead, it feeds one view to the fast, and the other to the slow network, and uses the cosine distance between the two outputs as loss.

SwAV~\cite{caron2020unsupervised} is an interesting technique that, instead of using instance-based pairwise positive/negative examples, creates pseudo-labels by clustering the representations online, batch by batch. The pretext task is assigning data-augmented views of the same training sample to the same cluster, with an equipartitioning constraint preventing the trivial solution of a single cluster.

In contrast to the techniques above, which use standard data augmentation techniques to create the views of the samples for contrastive learning, InfoMin~\cite{tian2020makes} \textit{learns} how to create the views, using a criterion of minimizing the mutual information between views. The motivation is creating a challenging but feasible pretext task for the model.

\subsection{Self-supervised Learning on Medical Tasks} 

Currently, there are two paths to follow when the matter is using self-supervised learning in medical applications. One is simply to use the same pretext task designed for general purpose computer vision or propose a slightly adapted version of such tasks that fit best into the current medical application. Early medical applications leveraged self-supervised pretext tasks of reconstructing distorted or damaged inputs~\cite{selfsup-retina-2018,CHEN2019101539,moris-2021-eye,boyd2021self}, such as image reconstruction in retinal images~\cite{selfsup-retina-2018}, context restoration in fetal MRI~\cite{CHEN2019101539}, or depth estimation in monocular endoscopy~\cite{liu2018self}. Zhou et al.~\cite{surpassimagenet}, working on X-ray images, employ a domain-general pretext task (the matching of data-augment views of instances of most methods of Sec.~\ref{sec:related-sslgeneral}), and uses stronger baselines: both supervised pre-training on ImageNet, and self-supervised pre-training with MoCo~\cite{moco}, still showcasing improvements in downstream image classification. Their technique, \textit{Comparing to Learn}, uses two networks, in knowledge-distillation teacher-student pair, for a momentum encoding scheme somewhat reminiscent of BYOL~\cite{byol}. MoCo pre-training appears to be widely used in the medical field, bringing superior performance to other medical applications compared to their supervised counterpart for COVID diagnosis~\cite{vu2021medaug,sriram2021covid}, and pleural effusion classification~\cite{chen2021momentum}. 

On the other hand, the second way to explore self-supervised is to leverage knowledge about the medical domain – by experience or any domain expert involved – and computer vision to design a custom-built pretext task for the target medical application. Suitable pretext tasks are crucial for learning predictive representations, motivating some works to evaluate whether domain-specific might improve self-supervised learning for medical images. For instance, Jamaludin et al.~\cite{jamaludin2017self} pre-train a Siamese Net with a contrastive loss in which the positive pairs are patches of spinal magnetic resonance images depicting the same vertebrae of a patient across exams, and the negative pairs are corresponding vertebrae in different patients. They found that the scheme improves the prediction of intervertebral disc degeneration. Wenjia et al.~\cite{wenjiamri2019} use a pretext in which the model has to predict the bounding boxes of anatomic features in heart magnetic resonance images, metadata ordinarily available in the DICOM files. They found improvements in the downstream task of segmenting the heart in the images. An issue with all those works is their choice of baseline, networks initialized with random weights, instead of stronger baselines such as models fine-tuned on ImageNet, or other schemes for self-supervision. 

Most close to our work of performing a systematic evaluation, Truong et al.~\cite{truong2021transferable} assess four medical classifications and three distinct self-supervised pre-training in similar training regimens that ours but lacks evaluation regarding out-of-distribution performances. Hosseinzadeh et al.~\cite{hosseinzadeh2021systematic} evaluate the performance of fourteen self-supervised ImageNet models to a diverse set of tasks in medical image classification and segmentation. Again, out-of-distribution and low-data performance evaluations remained uncovered.

\subsection{Self-supervised Learning on Skin Lesion Analysis} 

Wang et al.~\cite{WANG2021102428} employ a clustering pretext-task reminiscent of SwAV~\cite{caron2020unsupervised}, but accumulating samples from several small batches, and employing different clustering and losses. Since the downstream task is \textit{unsupervised} learning on the same clusters, although evaluated on the classes of ISIC 2018 Lesion Diagnostic Challenge, this work is in a gray zone between self-supervised and purely unsupervised learning. They found favorable results compared to other clustering techniques, but, not surprisingly, a large penalty compared to works that employ supervised fine-tuning. Segmentation tasks also benefits from self-supervision, such as in Li et al.~\cite{li2020multi}, and Wang et al.~\cite{9761620}, which both applied the self-supervised with color-based pretext tasks for segmenting skin lesions.   

Most related to our work, Azizi et al.~\cite{azizi2021big} performed a well-designed, systematic evaluation of SimCLR~\cite{chen2020simple}, for two medical tasks: skin-lesion analysis on a private dataset of $>450,000$ teledermatology clinical images, and X-rays on the publicly available CheXpert dataset. Contrasting SimCLR pre-training to two strong supervised pre-training baselines, they find it advantageous for the skin-lesion task, and similar for the X-rays task. Their study is complementary to ours, with two medical tasks, three encoder architectures, three pre-training datasets, and the evaluation of a novel pretext technique for exploiting multiple images of the same clinical case they call Multi-Instance Contrastive Learning. Our study, whose focal point is skin-lesion diagnosis, evaluates five test datasets with in- and out-of-distribution images, three pipelines for self-supervision pre-training (in contrast to a challenging supervised pipeline), and five candidate self-supervision schemes. 
Verdelho et al.~\cite{verdelho-isbi-2022} compare only two self-supervised learning approaches both quantitatively and qualitatively, but lacks low-data and out-of-distribution evaluation. 

We stand out our work from the ones available in the literature in Table~\ref{tab:distinctions} by highlighting the contributions of our experimental design.

\begin{table}[th]
\scriptsize
    \centering
    \caption{Overview of related works that evaluate self-supervised \textit{vs.} supervised pre-training.}
    \begin{tabular}{lrrrrr}
        \toprule
         \multirow{2}{*}{Work$_{year}$} & \#Evaluated & Out-of-distribution & Low-data \\
          & Methods &  Evaluation & Evaluation  \\
         \midrule
         
         Azizi et al.~\cite{azizi2021big}$_{2021}$ & 2 & No & Yes \\

         Hosseinzadeh et al.~\cite{hosseinzadeh2021systematic}$_{2021}$ & 15  & No & No \\
        
         Truong et al.~\cite{truong2021transferable}$_{2021}$ & 5 & No & Yes \\
         
         Verdelho et al.~\cite{verdelho-isbi-2022}$_{2022}$ & 2 & No & No \\
         
         Ours$_{2022}$ & 6 & Yes & Yes \\

         \bottomrule
    \end{tabular}
    \label{tab:distinctions}
\end{table}



%


\section{Materials and Methods}
\label{sec:methods}

This section details the methodology, comprising the datasets and factors in our experimental design. We also discuss how we conduct the experimental evaluation of all pipelines. 


\subsection{Datasets}
\label{sec:datasets}


Following ISIC 2020 Challenge~\cite{rotemberg2021patient}, our task is melanoma \textit{vs.} benign lesions classification. We evaluate our experiments in five, high-quality, publicly available datasets (Table~\ref{tab:testsets}). 

\begin{table*}[h]
    \centering
    \caption{Description of the datasets used in this work. Mel.: number of melanomas. $\dagger$Split used for test if omitted}
    \resizebox{\textwidth}{!}{%
    \begin{tabular}{lrr>{\raggedright}p{6.25cm}>{\raggedright\arraybackslash}p{5.5cm}}
        \toprule
         Dataset (split$\dagger$) & Size & Mel. & Lesion Diagnoses & Other Information  \\
         \midrule
         isic19~\cite{isic} (train) & $14\,805$ & $3121$ & Melanoma \textit{vs.} actinic keratosis, benign keratosis, dermatofibroma, melanocytic nevus, vascular lesion & Dermoscopic images. \vspace{.05cm}\\
         isic19 (validation) & $1\,931$ & $224$ & Idem & Dermoscopic images,  in-distribution.\vspace{.05cm}\\
         isic19 (test)& $3\,863$ & $396$ & Idem & Idem. \vspace{.05cm}\\
         isic20~\cite{rotemberg2021patient} & $1\,743$ & $581$ & Melanoma \textit{vs.} actinic keratosis, benign keratosis, lentigo, melanocytic nevus, unknown (benign) & Dermoscopic images, out-of-distribution, additional unknown diagnosis.\vspace{.05cm}\\
         derm7pt–derm~\cite{Kawahara2018-7pt} & $872$ & $252$ & Melanoma \textit{vs.} melanocytic nevus, seborrhoeic keratosis & Dermoscopic images, out-of-distribution. \vspace{.05cm}\\
         derm7pt–clinic~\cite{Kawahara2018-7pt} & $839$ & $248$ & Idem & Clinical images, out-of-distribution. \vspace{.05cm}\\
         pad-ufes-20~\cite{padufes20} & $1\,261$ & $52$ & Melanoma \textit{vs.} actinic keratosis, Bowen’s disease, nevus, seborrheic keratosis & Clinical images, out-of-distribution, additional Bowen's disease diagnosis. \vspace{.05cm}\\
         \bottomrule
    \end{tabular}%
    }
    \label{tab:testsets}
\end{table*}


We performed all training and validation in splits of the isic19 dataset. We removed samples from isic20 present in the isic19 train/validation splits to avoid contaminating the former. We removed basal cell carcinomas and squamous cell carcinomas from all datasets, leaving melanoma as the only malignant class.

The diversity of test datasets aimed at mitigating bias in evaluation~\cite{geirhos2020shortcut,bissoto2020debiasing,bissoto2019constructing}, providing both in-distribution (same dataset, same type of image, same~classes) and out-of-distribution (cross-dataset, different types of image, different classes) scenarios.




\begin{figure}[ht]
    \centering
    \includegraphics[width=0.5\columnwidth]{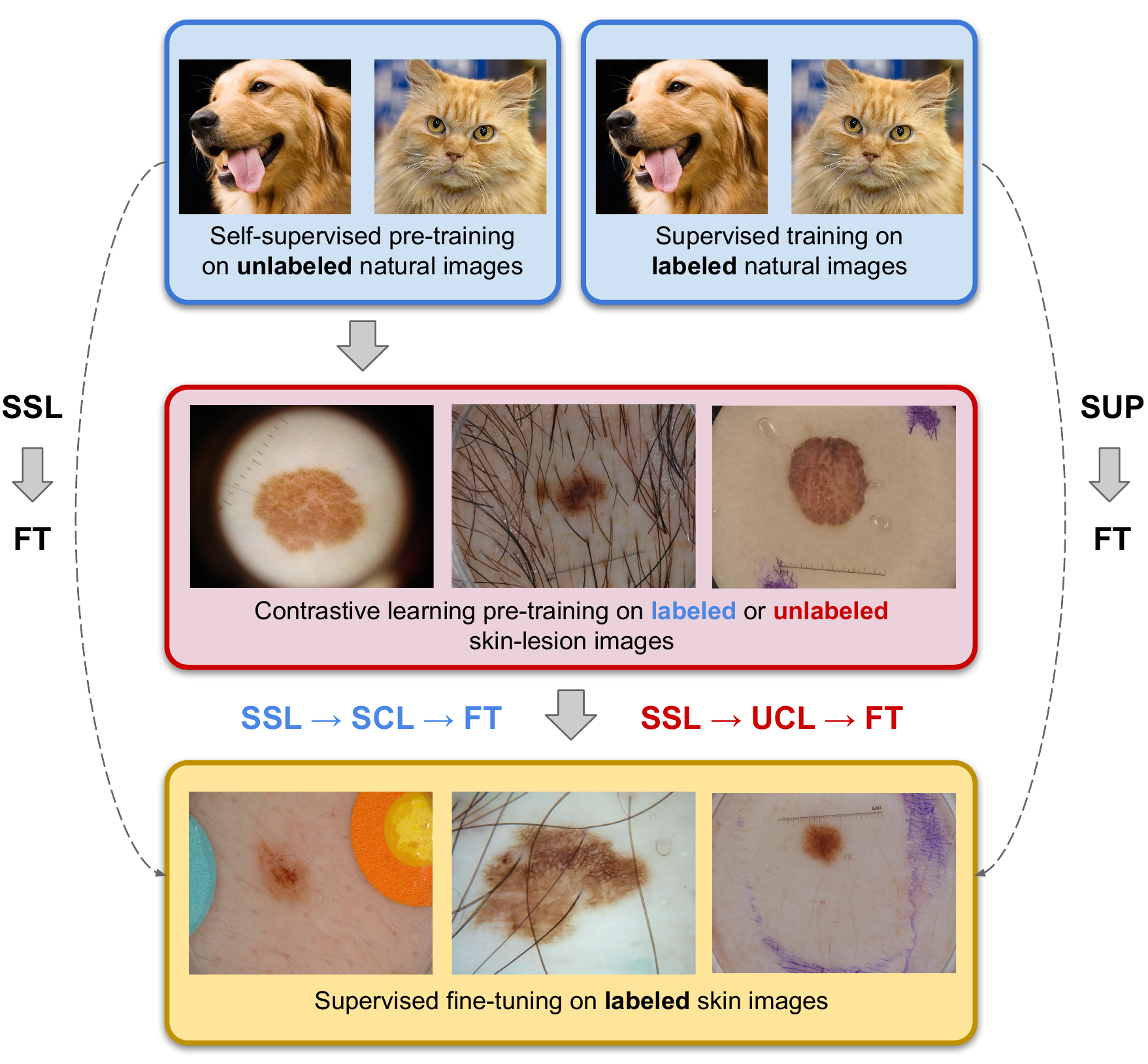}
    \caption{An overview of our evaluated pipelines. In SSL $\rightarrow$ FT scheme we contrast the result of five fine-tuned SSL ImageNet pre-trained models on isic19 dataset (see Sec.~\ref{sec:design}) with the supervised counterpart. The SSL $\rightarrow$ SCL $\rightarrow$ FT pipeline differs from SSL $\rightarrow$ UCL $\rightarrow$ FT according to the employed contrastive loss. They both go through a pre-training stage (see Sec.~\ref{sec:ssl-ucl-scl-ft}) --- which can be supervised (SCL) or unsupervised (UCL) --- using the isic19 dataset and then performing a supervised fine-tuning. Figure inspired from Azizi et al.~\cite{azizi2021big}.
    }
    \label{fig:all-protocols}
\end{figure}

\subsection{Experimental Design}
\label{sec:design}


We evaluate four alternative pipelines (Fig.~\ref{fig:all-protocols}), which vary in the pre-training and fine-tuning of the model. First, we wish to compare the baseline (supervised pre-training) pipeline (SUP $\rightarrow$ FT) with the basic self-supervision pipeline (SSL $\rightarrow$ FT) to establish whether self-supervision is advantageous. In addition, we wish to select a self-supervision scheme among five candidates (BYOL,  InfoMin, MoCo-V2, SimCLR, and SwAV) to perform the remainder of the experiments. Selecting the most promising scheme at this stage is necessary for managing the number of experiments, as the next round of experiments will be extensive and, thus, expensive. The \text{SSL $\rightarrow$ *  $\rightarrow$ FT} pipelines have an additional, intermediate pre-training step on the isic19 train split using supervised (SCL) or unsupervised (UCL) contrastive loss (Sec.~\ref{sec:ssl-ucl-scl-ft}). 


In the first round of experiments, we attempt a few combinations of hyperparameters for each self-supervision scheme. We purposefully optimize the baseline pipeline more thoroughly to make it challenging. The exact search space appears in Sec.~\ref{subsec:implementation-details}. We perform all searches on the isic19 validation split to avoid using privileged test information on this step~\cite{VALLE2020303}. To estimate the statistical variability of those experiments, we perform five replicates for every experiment, reflecting different random initializations for the training procedures (optimizer, scheduler, and augmentations). 

The next round of experiments is a systematic evaluation of all pipelines (Fig.~\ref{fig:all-protocols}) under three data regimens: full training data with 100\% of the samples, and low training data with 10, and 1\% of the samples. The latter intends to simulate the frequent scenario on medical images of insufficient training data.

For each combination of pipeline and hyperparameter (see Subsec.~\ref{subsec:implementation-details}), we measure their performance on the isic19 validation split five times, reflecting different random initializations for the training procedures, and, on the low-data experiments, also different random training subsets. We pick the five non-unique best combinations of hyperparameters for each pipeline. For each combination, we perform five replicates on the isic19 test split, resulting in 25 measurements for each pipeline. 








\subsection{SSL $\rightarrow$ UCL/SCL $\rightarrow$ FT Pipelines} 
\label{sec:ssl-ucl-scl-ft}

These two pipelines investigate the benefit of introducing an additional contrastive learning pre-training step before the traditional fine-tuning. Even though it adds an additional computational cost, works in the literature report some advantages~\cite{azizi2021big,azizi2022robust,cole2022does}, but their evaluation only consider domains with abundant data availability (compared to ours). To this end, we investigate if the same observed improvements also translate to domains with only a few hundred data available.  



We evaluated two contrastive losses:

\noindent\textbf{Unsupervised Contrastive Loss (UCL):} we performed the pre-training on the isic19 training set using the self-supervised NT-Xent contrastive loss~\cite{chen2020simple}:

\begin{equation}
\small
    \mathcal{L}_{UCL} = \frac{-1}{2N} \sum_{i=1}^{2N} \log 
        \frac{ \exp(z_i\cdot z_i^+) / \tau }
             { \sum_{k\neq{}i}^{2N} \exp(z_i\cdot z_k) / \tau },
    \label{eq:unsup-contrastive-loss}
\end{equation}

\noindent  where $z_{*}  = f(x_{*})$ is the representation for input $x_{i}$ output by the encoder $f$, $z^+_i$ is the positive pair to $z_i$, and all vectors $z_{*}$ are $\ell_2$-normalized. The scalar temperature $\tau$ hyperparameter regulates the concentration/spreading of the loss. Only the input data is exploited, and class labels are ignored.

\noindent\textbf{Supervised Contrastive Loss (SCL):} we performed pre-training using a straightforward extension of the loss above~\cite{khosla2020supervised}, which incorporates class labels by grouping as positive all examples in the same class (instead of just the augmented pair coming from the same instance):
\begin{equation}
\small
    \mathcal{L}_{SCL} = \frac{-1}{2N} \sum_{i=1}^{2N} \frac{1}{|Z_i^+|} \sum_{z^+\in{}Z_i^+} \log 
        \frac{ \exp(z_i\cdot z^+) / \tau }
             { \sum_{k\neq{}i}^{2N} \exp(z_i\cdot z_k) / \tau },
    \label{eq:sup-contrastive-loss}
\end{equation}
\noindent  where $Z^+_i$ is the set of all representations that are positive to $z_i$, and the other symbols are the same as in Eq.~\ref{eq:unsup-contrastive-loss}.




\subsection{Implementation Details}
\label{subsec:implementation-details}


We follow standard guidelines for self-supervised learning literature~\cite{goodpractices-ssl} and use ResNet-50 (1$\times$)~\cite{resnet} as base encoder for all experiments. In SUP $ \rightarrow $ FT scheme, we strive to make the baseline challenging, by performing, on the isic19 validation split, a thorough a grid search comprising batch size $(32, 128, 512)$, balanced batches (yes or no), starting learning rate $(0.1, 0.05, 0.005, 0.009, 0.0001)$, and learning rate scheduler (plateau, cosine). The optimizer is the SGD with a momentum of $0.9$ and weight decay of $0.001$. The plateau scheduler has patience of $10$ epochs and a reduction factor of $10$. 

For the SSL $ \rightarrow * \rightarrow$ FT pipelines, we employ two fully-connected layers to embed the ResNet-50 onto $128$-dimensional representations, fed to the contrastive loss. We resized the input images to $224\times224$ and used SimCLR's recommended heavy image augmentation pretexts — color jitter, horizontal and vertical flips, random resized crop, and grayscale. We omitted the Gaussian blur because, for skin lesions, it leads many images to be very similar, harming the results. We used a learning rate of $0.001$ with a cosine decay on an Adam optimizer. We perform a grid search through pre-training batch size (80 or 512), balanced batches (yes or no), temperature scale (0.1, 0.5, 1.0), and pre-training epochs (50 or 200). 

For all pipelines, the fine-tuning lasts for $100$ epochs with early stopping with patience of $22$ epochs, monitored on the validation loss. Both schedulers have a minimum learning rate of $10^{-5}$. All experiments ran in a single RTX 5000 GPU, except for the SSL $\rightarrow$ UCL/SCL $\rightarrow$ FT pipelines which required two Quadro RTX 8000 GPUs. The source code to reproduce our work in addition to detailed descriptions about the data is available on our public repository\footnote{\url{https://github.com/VirtualSpaceman/ssl-skin-lesions}}.






\section{Results}
\label{sec:results}

As explained in Sec.~\ref{sec:design}, we organized our extensive experimental design in two rounds, corresponding to the next two subsections. In a third subsection, 
we analyze the second round of experiments in a low training~data scenario. 

\subsection{Self-supervision Schemes \textit{vs.} Baseline Comparison}
\label{sec:experiments-schemes}


In this first round of experiments, we compared the baseline pipeline (SUP $\rightarrow$ FT) to the basic self-supervision pipeline  (SSL $\rightarrow$ FT) with five self-supervision schemes (BYOL, InfoMin, MoCo-V2, SimCLR, and SwAV). We optimized the baseline and self-supervised pipelines as explained in Sec.~\ref{subsec:implementation-details}.

The results (Table~\ref{table:ft-results}) show that, despite having no access to the labels during the pre-training, and being less thoroughly optimized during the final fine-tuning, the models with self-supervised pre-training are very competitive. Indeed, two of the pipelines (SimCLR and SwAV) had averages above the ones in the baseline.

\begin{table*}[thb]
\scriptsize
\caption{Results for the first round of experiments, comparing the supervised SUP $\rightarrow$ FT baseline to the basic SSL $\rightarrow$ FT pipeline with five SSL schemes. The metric is the AUC on the isic19 validation split. Despite the baseline using label information on pre-training, and being more thoroughly optimized, self-supervision pre-training is still very competitive with~it. 
}
\begin{center}
\begin{tabular}{lcrrrr}
\toprule
    \multirow{2}{*}{Method} & \multirow{2}{*}{AUC (\%)} & \multicolumn{4}{c}{Hyperparameters} \\
           &        & learning rate & ~batch size & batches & ~scheduler\\
\midrule
    Sup. baseline                   & 94.8 \scriptsize{$\pm$ 0.6} & $0.009$ & $128$ & balanced & plateau\\ 
    \textbf{SimCLR~\cite{chen2020simple}} & \textbf{95.6} \scriptsize{$\pm$ \textbf{0.3}} & $0.01$ & $32$ & unbalanced& plateau\\
    SwAV~\cite{caron2020unsupervised}     & 95.3 \scriptsize{$\pm$ 0.6} & $0.01$ & $32$ & unbalanced & plateau \\ 
    BYOL~\cite{byol}                      & 94.6 \scriptsize{$\pm$ 0.5} & $0.01$ & $32$ & unbalanced & plateau \\
    InfoMin~\cite{tian2020makes}          & 94.4 \scriptsize{$\pm$ 0.5} & $0.001$ & $32$ & unbalanced & plateau\\ 
    MoCo-V2~\cite{chen2020improved}       & 93.9 \scriptsize{$\pm$ 0.7} & $0.001$ & $32$ & unbalanced & plateau\\ 
    \bottomrule
\end{tabular}%
\label{table:ft-results}
\end{center}
\end{table*}

This first round of experiments intended to validate the applicability of self-supervised learning, and to select one self-supervised scheme for the expensive round of systematic evaluations in the next round. Thus, it comes with the important caveat that both optimization and evaluation were conducted in the~isic19 validation set. The second round of experiments will evaluate the ability of the pipelines to generalize performance in the rigorous setting of a held-out test set.

\subsection{Systematic Evaluation of Pipelines}
\label{sec:experiments-pipelines}

In the second round of experiments, we performed a systematic evaluation of the baseline pipeline, pre-trained with supervision (SUP $\rightarrow$ FT) against the three pipelines pre-trained with self-supervision (SSL  $\rightarrow$ FT, SSL  $\rightarrow$ UCL $\rightarrow$ FT, and SSL  $\rightarrow$ SCL $\rightarrow$ FT). In this round, we only evaluated SimCLR as the self-supervision scheme for several reasons: it showed the best performance in the preliminary experiments (Sec.~\ref{sec:experiments-schemes}), it allows introducing annotation information easily with a supervised contrastive loss, it had one hyperparameter less than SwAV to optimize (number of clusters), and the ablation studies in the original papers helped to decide on a range of reasonable values for the temperature~value. 



As explained in Sec.~\ref{sec:design}, this round of experiments simulates a realistic machine-learning protocol, in which first we optimize the hyperparameters for each pipeline on the isic19 validation split, then evaluate the performance on a held-out test set. The test sets considered as in-distribution is isic19 test split; and isic20, derm7pt-derm, derm7pt-clinic, and pad-ufes20 as out-of-distribution. Those cross-dataset evaluations are critical to evaluate how well the pipelines generalize to different classes, image acquisition techniques, or even to subtle dataset variations across institutions. 

The results appear in the topmost plot of Fig.~\ref{fig:boxplots-allmodels}, where each boxplot shows the distribution of 25 individual measurements (small black dots), corresponding to the best five non-unique hyperparameterizations, with five replicates for each of them. The boxplots show, as usual, the three quartiles (box), and the range of the data (whiskers) up to $1.5\times$ the interquartile range (samples outside that range are plotted individually as “outliers”). The large red dots show the means for each experiment. The metric is the AUC on the test datasets labeled on the right vertical axis. To make the horizontal axis comparable across its domain, we linearize the AUC using the logit (i.e., the logarithm of the odds) in base 2, shown on the bottom axis. The original AUC values appear on the top axis.

The plots reveal two advantages for the self-supervised pipelines: first, performances (means and medians) tend to be higher; second, the variability (width of the boxes) tended to be smaller. That shows the ability of the self-supervised pre-training in improving the results and in making them more~stable. 

No consistent advantage in terms of trend improvement (mean, median) is evident among the different self-supervised pipelines, but in terms of variability reduction, the double-pre-trained pipelines (SSL $\rightarrow$ SCL/UCL $\rightarrow$ FT) appear to have a slight advantage.

\subsection{Low Training Data Scenario}
\label{sec:experiments-low-data}

\begin{figure*}[phb]
    \centering 
    \includegraphics[width=\textwidth]{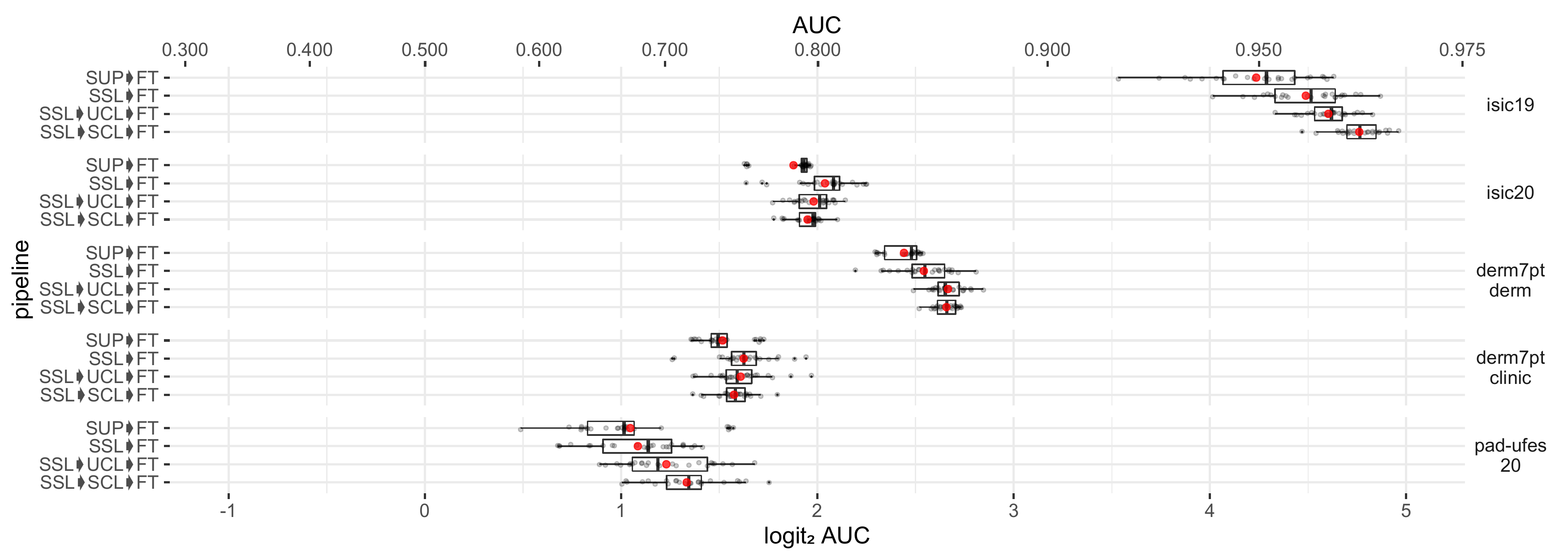}
    \includegraphics[width=\textwidth]{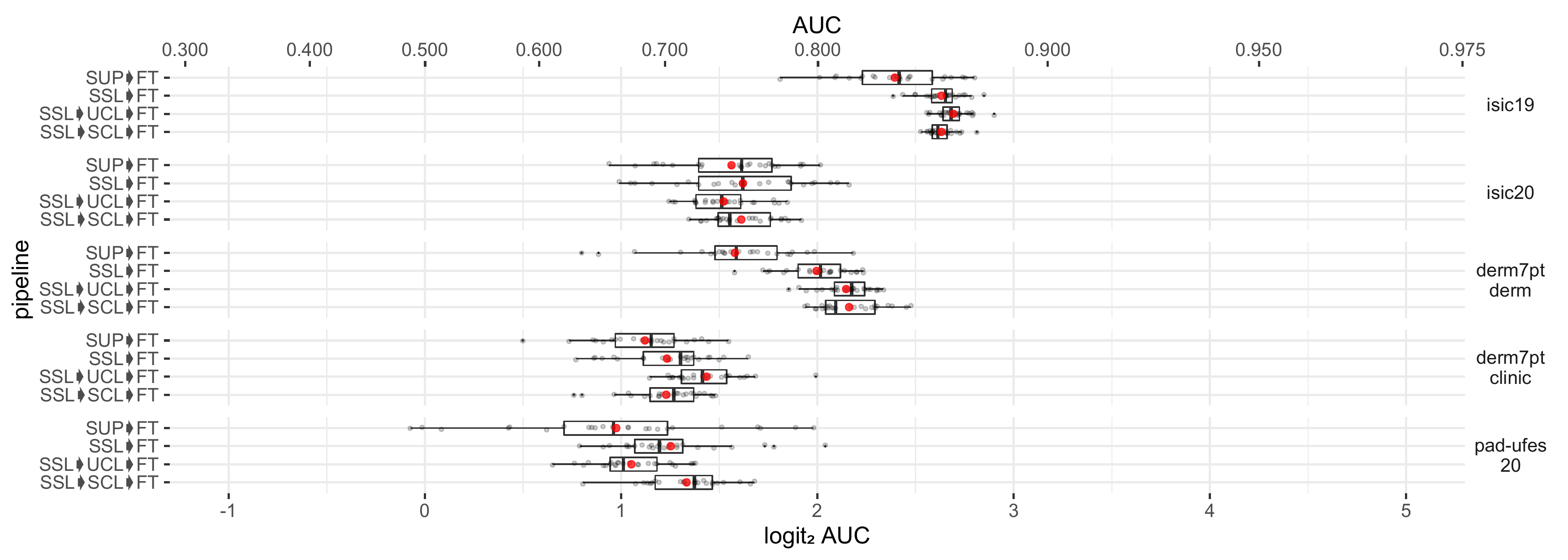}
    \includegraphics[width=\textwidth]{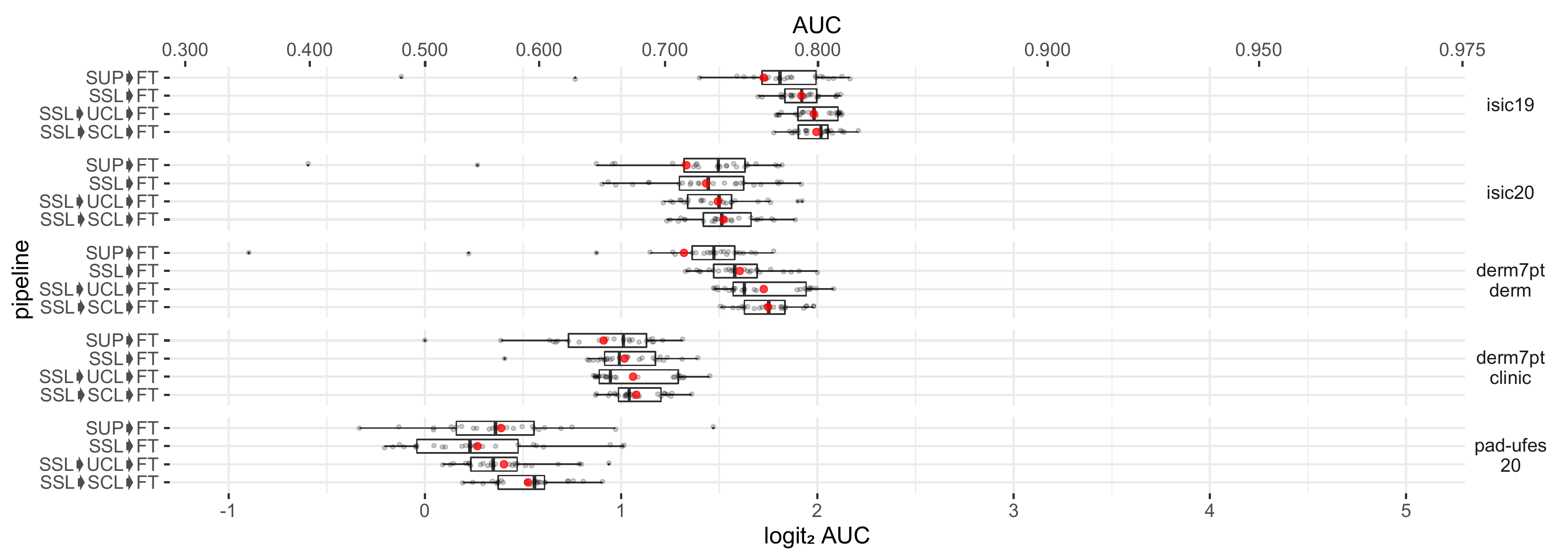}
    \caption{Results for the second round of experiments, with a systematic comparison of the pipelines labeled on the left vertical axis at the datasets labeled on the right vertical axis. The top, middle, and bottom plots show results for 100, 10 and, 1\% of the training data, respectively. Individual measurements represented by each boxplot appear as small black dots, whose means appear as larger red dots. In general, self-supervised pre-trained improves trends (medians, means) and reduces variability, both in the full-data and the low-data scenarios.}
    \label{fig:boxplots-allmodels}
\end{figure*}

These results follow the same protocol as those in the previous section but with drastically reduced train datasets. The results appear in the middle and bottom-most plots in Fig.~\ref{fig:boxplots-allmodels}, for 10\% (1480 samples) and 1\% (148 samples), respectively, of the original train dataset. Other than for this restriction, the interpretation of the plots is the same as in the previous section.

The results are much noisier than the full-data experiments: in part, this is intrinsic to the smaller training sets, but the random choice of training subsets also contributes to increased variability. 

Again, the self-supervised pipelines appear advantageous, both in terms of trend improvement (mean, median) and in terms of variability reduction, but here the advantage of the double-pre-trained pipelines (SSL $\rightarrow$ SCL/UCL $\rightarrow$ FT) seems more decisive, especially for the lowest data regimen, where it brings a clear improvement both in trend and variability. As we will discuss in the conclusions, such variability reduction is critical for the soundness of the deployment of low-data models.

\subsection{Qualitative Analysis}
\label{subsec:qualitative}

We performed a qualitative analysis, since we would like to have clues of what different pre-training methods learned to make the decision. Therefore, to analyze the differences between the learned presentations for each pre-training, we performed the 
Score-CAM~\cite{wang2020score}, 
method for visualizing the features learned by a neural network and 
the regions that activate a certain label.
We preferred a gradient-free method which overcome both saturation, and false confidence issues~\cite{wang2020score} from gradient-based techniques.  

\begin{figure}[t]
\captionsetup[subfigure]{justification=Centering}

\begin{tabular}{c c}
\rotatebox[origin=c]{90}{(a) True Positive}
\begin{subfigure}{0.95\textwidth}
    \includegraphics[width=\textwidth]{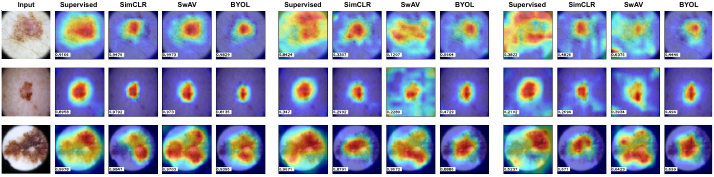}
\end{subfigure}
\end{tabular}

\medskip

\begin{tabular}{c c}
\rotatebox[origin=c]{90}{(b) False Positive}
\begin{subfigure}{0.95\textwidth}
    \includegraphics[width=\textwidth]{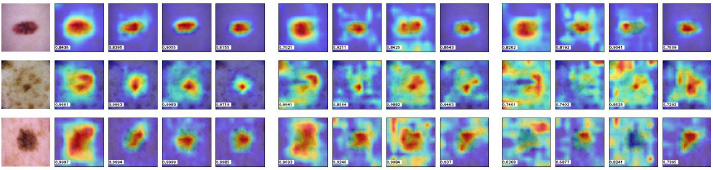}
\end{subfigure}
\end{tabular}

\medskip

\begin{tabular}{c c}
\rotatebox[origin=c]{90}{(c) True Negative}
\begin{subfigure}{0.95\textwidth}
    \includegraphics[width=\textwidth]{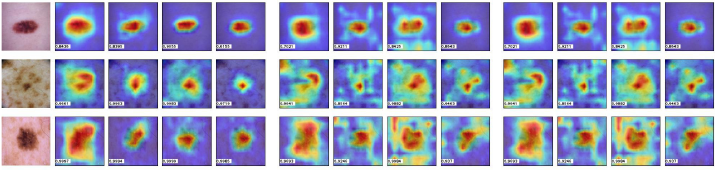}
\end{subfigure}
\end{tabular}

\medskip

\begin{tabular}{c c}
\rotatebox[origin=c]{90}{(d) False Negative}
\begin{subfigure}{0.95\textwidth}
    \includegraphics[width=\textwidth]{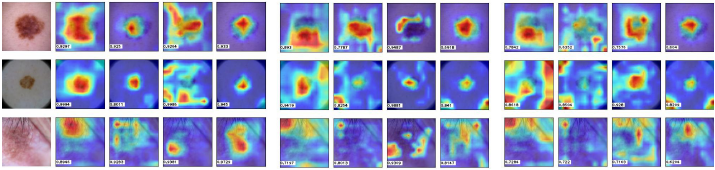}
\end{subfigure}
\end{tabular}

\caption{Results of Score-CAM visualization for the top-3 self-supervised models \textit{vs.} the supervised baseline for each training data regime. The brighter the more relevant the corresponding region in the image is for the model prediction. We randomly sampled three images from the isic19 test set to show true positive, false positive, true negative, and false negative scenarios; considering the full-data training case. For example, in a true positive scenario, we take three random images in which all images are classified as malignant correctly.}
\label{fig:mosaic}
\end{figure}

Fig.~\ref{fig:mosaic} shows the Score-CAM results for the top-3 self-supervised models (according to Table~\ref{table:ft-results}) \textit{vs.} the supervised baseline for each training regime (100\%, 10\%, and 1\% of training data). We randomly sampled three images from the isic19-test to show true positive, true negative, false positive, and false negatives cases. Apart from having different confidences about the prediction of each sample (bottom left rectangle) each model appears to highlight distinct regions. In general, both confidences and activation maps in true positive, and true negative seems to focus on similar regions in the 100\% training regime, but the attention appears sparser in the 10 and 1\% training data regime. 





\section{Conclusions}
\label{sec:conclusion}

Our experiments show that self-supervised pre-training makes models easier to deploy than classical supervised pre-training. Even with a less thorough hyperoptimization, the former outperformed the latter in general trends and — especially — in variability. 

It is hard to quantify this impression, but the models pre-trained with self-supervised also ``felt'' easier, faster, and more ``ready-to-use'' than the baseline models during training.

The advantage of the self-supervised pipelines was particularly prominent in the low-data scenarios, where their ability to stabilize the results, reducing variability, was even more noticeable. In those scenarios, especially the very-low data one, the double-pre-trained models appeared advantageous. We conjecture that self-supervised pre-training improved the mean performance due to the task granularity~\cite{cole2022does,azizi2022robust}, in which self-supervised models showed to perform particularly better when the pre-training is empowered by contrastive learning. However, those findings are shown experimentally, without any theoretical formulation. Understanding what circumstances make self-supervised competitive (or even superior) from a theoretical perspective is a promising research area. 

Very low-data scenarios are not, unfortunately, rare in medical applications. Models trained on such regimens will experience large variances in performance in contrast to models trained with adequate samples, but model designers will often be unaware of such variance (since they cannot run a simulation such as ours, comparing their model to others trained in different datasets of the same size). Our results suggest that self-supervised pre-training may reduce that variability, leading to saner models. Of course, models trained on small samples may suffer from severe biases, and extensive exploration is necessary to evaluate whether self-supervised might reinforce those biases~\cite{bissoto2019constructing}. In addition, we performed qualitative analysis using Score-CAM to visualize the network's attention. 

Self-supervised learning is a thriving research area, and the possibility of creating domain-specific or at least domain-aware pretext tasks for skin-lesion analysis is an exciting avenue of continuation for this work. Domain-aware sampling methods for selecting the positive and negative pairs in contrastive learning — even while using commodity pretext tasks — also instigate the possibility of incorporating domain knowledge into self-supervision learning.

\section*{Acknowledgements}

L. Chaves is partially funded by Santander and Google LARA 2021. 
A. Bissoto is funded by FAPESP 2019/19619-7. 
E. Valle is partially funded by CNPq 315168/2020-0. 
S. Avila is partially funded by CNPq 315231/2020-3, FAPESP 2013/08293-7, 2020/09838-0, and Google LARA 2021. This study was financed in part by the Coordenação de Aperfeiçoamento de Pessoal de Nível Superior -- Brasil (CAPES) -- Finance Code 001. The Recod.ai lab is supported by projects from FAPESP, CNPq, and CAPES.

\clearpage
%
%
\bibliographystyle{splncs04}
\bibliography{egbib}
\end{document}